\documentclass{article}

\usepackage{sty/packages}
\usepackage{sty/article}
\usepackage{sty/bibliography}
\usepackage{sty/macros}

\newcommand{\cellname}{bistable memory recurrent unit}

\newcommand{\cell}{BMRU}

\renewcommand{\tanh}[1]{\text{tanh} \lpar #1 \rpar}

\newcommand{\cand}{\hat{h}}
\newcommand{\cd}{\circledast}

\title{Parallelizable memory recurrent units}

\author[1,2]{Florent De Geeter}
\author[1]{Gaspard Lambrechts}
\author[1]{Damien Ernst}
\author[1]{Guillaume Drion}

\affil[1]{\normalsize Montefiore Institute, University of Liege, Liege, Belgium}
\affil[2]{\texttt{\normalsize florent.degeeter@uliege.be}}

\date{}

\begin{document}

\maketitle

\begin{abstract}
    With the emergence of massively parallel processing units, parallelization has become a desirable property for new sequence models. The ability to parallelize the processing of sequences with respect to the sequence length during training is one of the main factors behind the uprising of the \textit{Transformer} architecture.
    However, Transformers lack efficiency at sequence generation, as they need to reprocess all past timesteps at every generation step. Recently, \textit{state-space models} (SSMs) emerged as a more efficient alternative. These new kinds of recurrent neural networks (RNNs) keep the efficient update of the RNNs while gaining parallelization by getting rid of nonlinear dynamics (or recurrence). SSMs can reach state-of-the-art performance through the efficient training of potentially very large networks, but still suffer from limited representation capabilities. In particular, SSMs cannot exhibit persistent memory, or the capacity of retaining information for an infinite duration, because of their monostability. In this paper, we introduce a new family of RNNs, the \textit{memory recurrent units} (MRUs), that combine the persistent memory capabilities of nonlinear RNNs with the parallelizable computations of SSMs. These units leverage multistability as a source of persistent memory, while getting rid of transient dynamics for efficient computations. We then derive a specific implementation as proof-of-concept: the \cellname{} (\cell{}). This new RNN is compatible with the parallel scan algorithm. We show that \cell{} achieves good results in tasks with long-term dependencies, and can be combined with SSMs to create parallelizable hybrid networks that have both transient dynamics and persistent memory.
\end{abstract}

\section{Introduction}
\label{sec:introduction}

Recurrent neural networks (RNNs) \citep{hochreiter1997Long,cho2014Learning} used to be the most popular architecture for tackling sequential tasks thanks to their ability to model a large range of dynamical systems, until being foreshadowed by the Transformer architecture \citep{vaswani2017Attention}. The impossibility to parallelize RNNs with respect to time for training made Transformers a much suited architecture for building large models and training them on large amount of data. However, for generative modeling tasks, the Transformer suffers from a high computational complexity. Indeed, at every timestep of generation, it needs to process the full sequence of past inputs. RNNs have however regained interest in recent years thanks to the success of state-space models (SSMs) \citep{gu2022Efficiently}. SSMs use strictly linear dynamics in cascade with static nonlinearities, making their computation easily parallelizable. This allows SSMs to be used in large networks and to be trained on large datasets, because they can benefit from the computational power offered by GPUs. SSMs can compete with Transformers on a variety of tasks \citep{gu2022Efficiently}, and large-language models based on SSMs like Mamba \citep{gu2024Mamba} have achieved comparable or better performance than Transformers for similar model sizes. SSMs are more efficient than Transformers during inference, as their recurrent formulation allows them to compute each output only from their previous output and the input, while Transformers require the whole sequence of inputs to be fed (or cached activations along the whole sequence of inputs).
The strictly linear recurrent dynamics of SSMs however creates some limitations: SSMs can only approximate fading-memory systems, i.e. systems that have a unique stable state \citep{boyd1985Fading}. Monostable systems such as SSMs encode past information in their transient dynamics, which creates a memory that ineluctably fades over time and makes some tasks unsolvable for a fixed number of layers \citep{merrill2024Illusion}.
Alternatively, multistable systems have several stable states, which allows to encode information for an infinite duration. It is possible to foster multistability in RNNs \citep{vecoven2021bioinspired,lambrechts2023Warming}, and multistability has been shown to drastically improve RNN performance especially in long-term dependencies tasks \citep{lambrechts2023Warming}. Multistability however requires strongly nonlinear recurrent connections, which usually prevents parallelization. In this configuration, multistable RNNs cannot compete with parallelizable alternatives such as SSMs or Transformers.

This work attempts to close this gap through the creation of novel RNN models that are parallelizable over the sequence length but exhibit persistent memory through multistability. \Cref{sec:rnns} first introduces the concept of RNNs with internal clocks, i.e. RNNs in which internal states can be updated several times between two consecutive timesteps, as a way to formalize our design approach. \Cref{sec:convergent_rnns} focuses on the RNN class where infinite updates can occur between two consecutive timesteps. From that, it introduces the concept of \textit{memory recurrent unit} (MRU), a class of RNNs that exhibit persistent memory but no fading memory. \Cref{sec:cell} further introduces \textit{\cellname{}} (\cell{}), a concrete example of MRU that can be trained on sequential tasks. \cell{} properties and performance are finally evaluated on several benchmarks in \cref{sec:experiments}.

\section{RNNs with internal clocks}
\label{sec:rnns}
RNNs encode time-dependencies in a recurrent hidden state $h_t$ whose update depends on the previous hidden state $h_{t-1}$ and current observation of a time-series $x_t$. It writes
\begin{equation}
    h_t = f_{\theta} \lpar x_t, h_{t-1}; \theta \rpar, \forall t \geq 1 \label{eq:RNN}
\end{equation}
where $f_{\theta}$ represents the update equation of the RNN and $\theta$ the parameters of the network. The key mechanism is the recurrent connection that defines a dependency between $h_t$ and $h_{t-1}$. Nonlinear RNNs such as LSTM \citep{hochreiter1997Long} and GRU \citep{cho2014Learning} have remained state-of-the-art methods in sequence modeling for several decades.

The classical RNN formulation of \cref{eq:RNN} constraints the internal state $h_t$ to only be updated when the network receives a new input $x_t$, i.e. only once every timestep $t$. This constraint forbids the network to modify its internal state evolution between two timesteps. It particularly impedes our goal to quickly encode information on stable equilibria at convergence.

To release this constraint, we define a class of RNNs whose internal dynamics are decoupled from the external dynamics of the input time-series. Each RNN of this class has internal clocks that permit to update their internal states $N$-times between two timesteps $t$, where $N$ can be different for each unit. The internal update equation of such RNN unit between two timesteps writes
\begin{subequations}
    \begin{align}
        \Tilde{h}_t[0]   & = h_{t-1},                                                                        \\
        \Tilde{h}_{t}[n] & = f_{\theta} (x_t,\Tilde{h}_{t}[n-1]; \theta), \forall n \geq 1, \label{eq:clock} \\
        h_t              & = \Tilde{h}_t[N],
    \end{align}
\end{subequations}
where $t$ is the current timestep, $n \in [0, \dots, N]$ the current internal clock iteration and $N \in \N$ the number of internal clock iterations. $\Tilde{h}_t[n]$ is the transient internal state value at iteration $n$ and $h_t$ the final internal state at time $t$.

Classical RNNs are contained within this class. Indeed if we set $N=1$, \cref{eq:clock} becomes
\begin{equation*}
    \Tilde{h}_{t}[1]     = f_{\theta} (x_t,\Tilde{h}_{t}[0]; \theta)
\end{equation*}
with $\Tilde{h}_t[0] = h_{t-1}$ and $\Tilde{h}_t[1] = h_t$, which leads to \cref{eq:RNN}.

For $1<N<\infty$, we have the set of RNNs whose internal dynamics are decoupled from the time-series dynamics, as their internal state is updated $N$ times between two timesteps. Such RNNs can be trained as classical RNNs, but the potential increase in expressive power comes at a computational cost, as the effective sequence length becomes $N\cdot T$ at inference, $T$ being the length of the time-series. Using linear recurrence of the form
\begin{equation*}
    f_{\theta} (x_t,\Tilde{h}_{t}[n]; \theta) = A(\theta)\Tilde{h}_{t}[n-1] + B(\theta)x_t,
\end{equation*}
can mitigate this complexity, as the value $\tilde{h}[N]$ can be computed analytically from $\tilde{h}[0]$ while maintaining the possibility to output potentially complex trajectories. This is for instance the approach taken by practical implementations of state-space models (SSMs)\citep{smith2022Simplified}.

For $N\to\infty$, we have the specific set of ``convergent'' RNNs, i.e. RNNs that can converge towards their steady-state values between two timesteps. In this work, we use the framework of convergent RNNs to build a novel type of RNN unit that uses stable states as the basis for their computations.

\section{The specific case of convergent RNNs (\texorpdfstring{$N\to\infty$}{N→∞})}
\label{sec:convergent_rnns}

We focus on RNNs that converge towards stable equilibria, i.e. RNNs whose update equation $f_{\theta}(x_t, h_{t-1};\theta)$ is such that:
\begin{equation*}
    \forall x_t, \tilde{h}_t[0], \; \exists h^* \in \R : \lim_{N \to \infty} f_{\theta}^N(x_t, \tilde{h}_t[0]; \theta) = h^*,
\end{equation*}
where:
\begin{equation*}
    f_{\theta}^n(x_t, \tilde{h}_t[0]; \theta) = \begin{cases}
        f_{\theta}(x_t, f_{\theta}^{n-1}(x, \tilde{h}_t[0]; \theta); \theta) & \text{ if } n > 1, \\
        f_{\theta}(x_t, \tilde{h}_t[0]; \theta)                              & \text{ if } n = 1.
    \end{cases}
\end{equation*}

Letting the internal clock to loop an infinite number of times ($N\to\infty$) ensures that the RNN has converged towards a stable equilibrium between two time steps. This property, which is specific to this class of RNNs, ensures a stationary output, i.e. that the output remains stable over time if the input barely changes. This property is of great interest, as it ensures that the information encoded in the RNN unit will not fade over time. It however requires specific characteristics to be exploitable in practice.

\begin{figure}[tbp]
    \centering
    \includegraphics{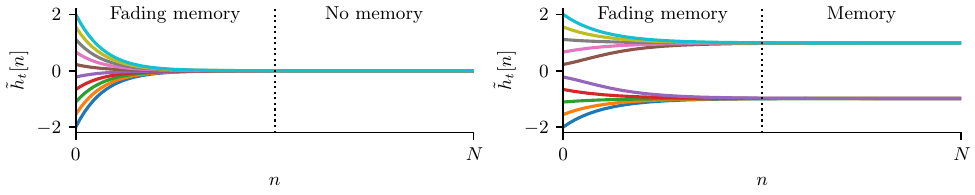}
    \caption{\textbf{Monostability vs bistability in a RNN with internal clock}. The figure shows internal state trajectories of the RNN unit described by \cref{eq:bistRNN} for different initial conditions $\tilde{h}_t[0] = h_{t-1}$. (\textbf{left}) Evolution of the system when $\beta = -1.5$. (\textbf{right}) Evolution of the system when $\beta = 1.5$.}
    \label{fig:fadingvspersistent}
\end{figure}

To illustrate this, we consider the following RNN update equation:
\begin{align}
    h_t = (1-c)\cdot h_{t-1} + c\cdot\tanh{x_t + (\beta + 1) \cdot h_{t-1}}, \forall t \geq 1,\label{eq:bistRNN}
\end{align}
which is a simplified version of the BRC and nBRC derived by \citet{vecoven2021bioinspired}. To construct the convergent version, we rewrite \cref{eq:bistRNN} at timestep $t$ with an internal clock
\begin{subequations}
    \begin{align}
        \Tilde{h}_t[0] & = h_{t-1}                                                                                                                           \\
        \Tilde{h}_t[n] & = (1-c)\cdot \Tilde{h}_{t}[n-1] + c\cdot\tanh{x_t + (\beta + 1) \cdot \Tilde{h}_{t}[n-1]}, \forall n \geq 1\label{eq:bistRNN_clock} \\
        h_t            & = \Tilde{h}_{t}[N]
    \end{align}
\end{subequations}
and simulate its response over several internal clock iterations $N$ for different values of previous state value $h_{t-1}$ and two different values of the parameter $\beta$ (\cref{fig:fadingvspersistent}) (other parameter values are $c=0.01$ and $x_t=0$). In both simulations, the trajectories corresponding to different values of $h_{t-1}$ are transiently distinct, until they all converge towards a steady-state. The distinct transient trajectories create fading memory, which can encode quantitative information about $h_{t-1}$ but only for a strictly finite amount of time.

Here, we focus our interest on what happens at convergence. In the case of $\beta=-1.5$ (\cref{fig:fadingvspersistent}, left), all initial $h_{t-1}$ lead to the same equilibrium point, and the RNN unit cannot maintain any stationary information about $h_{t-1}$. It is restricted to fading memory. This is because \cref{eq:bistRNN} for $\beta=-1.5$ leads to a monostable system, i.e. a system that has only one stable equilibrium for all input values $x_t$. Monostable RNNs are incapable of encoding stationary (i.e. persistent) information about the past. It includes all RNNs with linear recurrence, such as SSMs, as well as all gated RNNs in which gates do not depend on past state values and whose update gate $z_t$ is strictly larger than 0 (for the formulation $(1-z_t) \odot h_{t-1}$) or smaller than 1 (for the formulation $z_t \odot h_{t-1}$)\citep{martin2018Parallelizing,feng2024Were, qin2023Hierarchically}.

In the case of $\beta=1.5$ (\cref{fig:fadingvspersistent}, right), the trajectories converge towards two different equilibria depending on $h_{t-1}$: $h_t \simeq +1$ for positive values of $h_{t-1}$ and $h_t \simeq -1$ for negative values of $h_{t-1}$. The RNN unit maintains a stationary, qualitative information about $h_{t-1}$. In order words, some information about $h_{t-1}$ is encoded in \textit{persistent memory}. This is because \cref{eq:bistRNN} for $\beta=1.5$ leads to a bistable system, i.e. a system that has two stable equilibria for a set of input values $x_t$.

Constructing a multistable RNN with infinite internal clock cycles would therefore permit to isolate its persistent memory capabilities. But constructing such RNN by looping a large number of times in \cref{eq:clock} would drastically slow down inference and breach the convergence guarantee. We rather exploit the fact that the internal state converges towards an equilibria at $N\to\infty$, i.e. $\Tilde{h}_{t}[n] = \Tilde{h}_{t}[n-1] = h_t$. To do so, we rewrite an equivalent formulation for \cref{eq:bistRNN_clock} at the convergence point
\begin{align*}
    \Tilde{h}_t[0] & = h_{t-1}                                                        \\
    h_t            & = (1-c)\cdot h_{t} + c\cdot\tanh{x_t + (\beta + 1) \cdot h_{t}},
\end{align*}
which can be rewritten
\begin{align}
    F(h_t, x_t; \beta) = 0; \quad \Tilde{h}_t[0] = h_{t-1}\label{eq:hyst}
\end{align}
where $F(h_t, x_t; \beta) = h_t - \tanh{ x_t + (\beta + 1) \cdot h_t }$.
\Cref{eq:hyst} provides a set of necessary conditions that the RNN unit must satisfy at convergence. This equation defines an implicit function that can have multiple solutions $h_t$ for a given input $x_t^*$. The solution that corresponds to the output at time $t$ depends on the value of $h_{t-1}$, which dictates towards which stable point $h_t^*$ the cell will converge (i.e., $h_{t-1}$ determines in which basin of attraction the trajectory lies). This solution must also satisfy the stability property to ensure that it is a stable point:
\begin{align*}
    \left.\frac{d f(x_t,h_t;\beta)}{d h_t} \right|_{x^*_t,h^*_t} < 1
\end{align*}
where $f(x_t,h_t;\beta) = (1-c)\cdot h_{t} + c\cdot\tanh{x_t + (\beta + 1) \cdot h_{t}}$.

\begin{figure}[t]
    \centering
    \begin{subfigure}{\textwidth}
        \centering
        \includegraphics{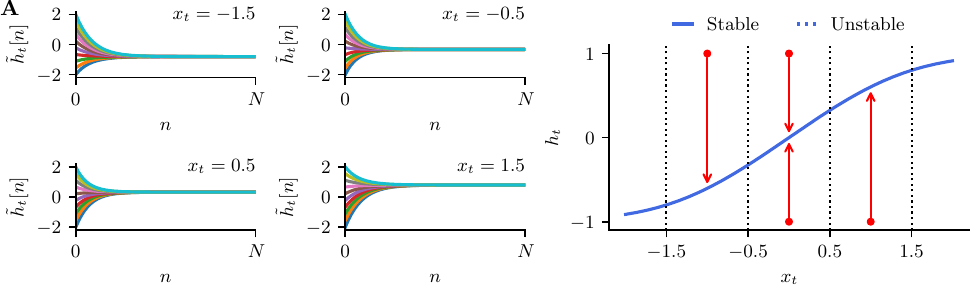}
    \end{subfigure}
    \begin{subfigure}{\textwidth}
        \centering
        \includegraphics{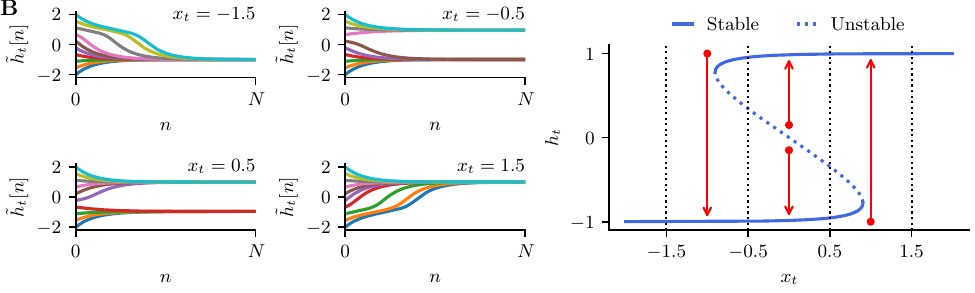}
    \end{subfigure}
    \caption{\textbf{Convergence properties of the RNN unit described by \cref{eq:bistRNN_clock} for different values of input $x_t$ and either $\beta=-1.5$ (\textbf{A}) or $\beta=1.5$ (\textbf{B}) } (\textbf{left}) Internal state trajectories corresponding to different initial conditions $\tilde{h}_t[0]=h_{t-1}$ and 4 different input values $x_t$. (\textbf{right}) Solutions of the steady-state equation \cref{eq:hyst} for $\beta=-1.5$ (\textbf{A}) and $\beta=1.5$ (\textbf{B}). The red arrows show convergence trajectories from different initial conditions $\tilde{h}_t[0]$.}
    \label{fig:steady_behavior}
\end{figure}

The solutions of \cref{eq:hyst} and their stability are shown in \cref{fig:steady_behavior} for $\beta=-1.5$ (A) and $\beta=1.5$ (B). For $\beta=-1.5$, the function has only one solution $h_t$ for all $x_t$, and the convergent RNN has a similar input-output function as a static layer. For $\beta=1.5$, the function has a memory region where two solutions are possible, and the selected solution $h_t$ depends on $h_{t-1}$. The RNN has encoded some information of the past input in persistent memory.

More generally, the set of necessary conditions that any RNN defined by \cref{eq:clock} must satisfy at convergence are determined by the implicit function
\begin{align}
    F_\theta(x_t,h_{t}; \theta) = 0, \label{eq:Ninf}
\end{align}
where $F_\theta(x_t,h_{t}; \theta) = h_{t}-f_{\theta}(x_t,h_{t}; \theta)$. It is therefore possible to efficiently compute the outputs of a convergent RNN at each timestep $t$ by using this implicit function as the update function
\begin{align}
    F_\theta(x_t,h_{t}; \theta) = 0, \forall t=1,...,T, \label{eq:gen_implicit}
\end{align}
where $h_{t-1}$ is used to select the output value when multiple solutions exist. This leads to a recurrent unit that only encodes information in persistent memory. We call this type of recurrent unit a \textbf{memory recurrent unit (MRU)}. It is important to note that \cref{eq:gen_implicit} can be solved in parallel for each timestep, as the function only depends on $x_t$ and $h_t$, but requires $h_{t-1}$ to select the proper solution based on history.

In dynamical systems theory, the implicit function $F_\theta(x_t,h_{t}; \theta) = 0$ corresponds to the bifurcation diagram of the system described by \cref{eq:clock}. For instance, \cref{eq:hyst} and associated \cref{fig:steady_behavior} correspond to a hysteresis bifurcation. Hysteresis bifurcations underlie bistability in existing RNN cells such as the bistable recurrent cell (BRC) \citep{vecoven2021bioinspired} (see \cref{app:Hyst_BRC}). More generally, a MRU can be built from any nonlinear dynamics by using its bifurcation diagram as an implicit activation function. The output state in the memory region can be computed using iterative methods such as the Newton–Raphson method \citep{lim2023Parallelizing,gonzalez2024Scalable,danieli2025ParaRNN}. Likewise, implicit differentiation can be used to compute exact gradients during backpropagation. The use of iterative methods can however slow down inference and hinder parallelization. In the next section, we show how this drawback can be circumvented, focusing on the hysteresis bifurcation and its associated bistability.

\section{A bistable memory recurrent unit}
\label{sec:cell}

\subsection{A computationally efficient hysteresis-based MRU}
Instead of directly using the implicit function derived by the hysteresis bifurcation diagram, we propose to create a MRU based on an approximation of this function that maintains all its qualitative properties, but which is computationally efficient and remain parallelizable. Indeed, unlike the general concept of MRU, this approximation does not require any iterative method as it is formulated as an explicit function that depends on the current input and the previous state. The goal of this section is to present this approximation, and to show how we build a \cellname{} (\cell{}) around it.

\begin{figure}
    \centering
    \begin{minipage}[t]{.49\textwidth}
        \centering
        \includegraphics{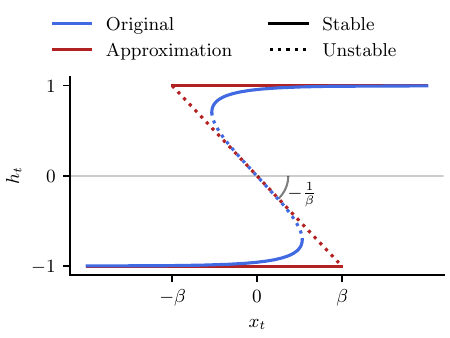}
        \caption{\textbf{Comparison between the implicit function and its approximation.} This figure compares the solutions $(h_t, x_t)$ of the implicit function defined by \cref{eq:hyst} (in blue) with the approximation defined by \cref{eq:approx} (in red) where $\alpha = 1$. Solid lines correspond to stable points, and dashed lines to unstable points.}
        \label{fig:implicit_vs_approx}
    \end{minipage}
    \hfill
    \begin{minipage}[t]{.49\textwidth}
        \centering
        \includegraphics{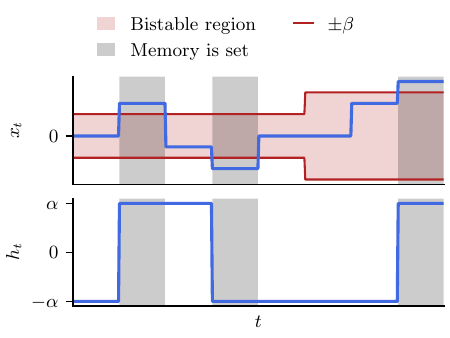}
        \caption{\textbf{Example simulation of the cell defined by \cref{eq:final_approx} for different values of $x_t$ and $\beta$.} The top graph defines two illustrative variations of $x_t$ and $\beta$, as well as the bistable region defined by $\beta$. The bottom graph shows the effect of variations in $x_t$ and $\beta$ on the evolution of the state $h_t$. The gray areas highlight the timesteps at which the memory is updated, which occurs when $x_t$ is outside of the bistable region.}
        \label{fig:approx}
    \end{minipage}
\end{figure}

We focus our design on the multistable behavior of the implicit function, i.e. $\beta > 0$, as $\beta \leq 0$ leads to a monostable, memory-less function (see \cref{fig:steady_behavior}). Under this condition, the original function can be decomposed into three parts: the upper stable points, the lower stable points, and the unstable points that serve as a boundary in the bistable region. We approximate these three parts by modeling the stable points as constant values $\pm\alpha$, and the \textit{boundary} function as a linear function whose slope is the slope of the original function at $(0, 0)$, i.e. $-\frac{1}{\beta}$.
The approximation writes
\begin{equation}
    h_t = \begin{cases}
        \alpha \cdot S(x_t)                         & \text{if } |x_t| \geq \beta, \\
        \alpha \cdot S(h_{t-1} + \frac{x_t}{\beta}) & \text{if } |x_t| < \beta,    \\
    \end{cases} \label{eq:approx}
\end{equation}
where $S$ denotes the \textit{sign} function (with $S(0) = 1$). Similarly to the original function, $h_t$ is independent of $h_{t-1}$ for a large input, but not for a small input. A comparison between this approximation and the original function is shown in \cref{fig:implicit_vs_approx}.

The convergence condition ensures that $h_{t-1} = \pm \alpha$ for all timesteps, which permits to further simplify \cref{eq:approx} for $|x_t| < \beta$:
\begin{equation}
    h_t = \begin{cases}
        \alpha \cdot S(x_t) & \text{if } |x_t| \geq \beta, \\
        h_{t-1}             & \text{if } |x_t| < \beta.    \\
    \end{cases} \label{eq:final_approx}
\end{equation}

It is only every time $|x_t|$ is greater than $\beta$ that $h_t$ is updated. Otherwise, it will remain constant. This property is illustrated in \cref{fig:approx}.

\subsection{A learnable \cellname{}}
\label{sec:a_learnable_bmru}
\Cref{eq:final_approx} defines the input-output properties of the BMRU, but does not contain any learnable parameters. Here, we add such parameters by taking inspiration from gated RNN structure.

First, we observe that the function defined by \cref{eq:final_approx} outputs two different values depending on the comparison between the values of $|x_t|$ and $\beta$. We can implement this function by introducing a \textit{binary} gate, $z_t$, that computes this condition, and rewrite \cref{eq:final_approx} as a gate-dependent update rule. It writes
\begin{subequations}
    \begin{align}
        z_t & = H(|x_t| - \beta), \label{eq:cell_zbis}
        \\
        h_t & = z_t \cdot S(x_t) \cdot \alpha + (1 - z_t) \cdot  h_{t-1}, \label{eq:cell_hbis}
    \end{align}
\end{subequations}
where $t$ is the current timestep, $H$ the Heaviside function and $S$ the sign function. \Cref{eq:cell_zbis} computes the binary gate value $z_t$, which is used to distinguish whether we are inside the bistable region ($z_t = 0$) or outside of it ($z_t = 1$). \Cref{eq:cell_hbis} then computes the new state value $h_t$ depending on the value of $z_t$.

We can then extend this formulation to the multidimensional case, where multiple inputs $x_t$ converge at a layer composed of multiple BMRUs. To be coherent with the variables used in gated RNNs, we name the input values $x_t$ and their combination at each BMRU cell $\hat{h}_t$, the \textit{candidate}. $x_t$ becomes a $M$-dimensional vector where $M$ is the number of inputs to the layer. Moreover, $\beta$ can be made input-dependent, and is therefore renamed $\beta_t$. $\hat{h_t}$, $\beta_t$ and $h_t$ become $N$-dimensional vectors, where $N$ is the number of BMRU cells in the layer. We use a classical fully connected layer to compute the vectors $\hat{h}_t$ and $\beta_t$ from the input vector $x_t$, adding the positivity constraint for $\beta_t$. The equations of the multidimensional BMRU network write
\begin{subequations}
    \begin{align}
        \cand_t & = W_x x_t + b_x, \label{eq:cell_cand}
        \\
        \beta_t & = |W_{\beta} x_t + b_{\beta}|, \label{eq:cell_beta}
        \\
        z_t     & = H(|\cand_t| - \beta_t), \label{eq:cell_z}
        \\
        h_t     & = z_t \odot S(\cand_t) \odot \alpha + (1 - z_t) \odot h_{t-1}, \label{eq:cell_h}
    \end{align}
\end{subequations}
where $\odot$ is the hadamard product, $t$ is the current timestep, $W_x$ and $W_{\beta}$ are matrices of learnable parameters, $b_x$, $b_{\beta}$ and $\alpha$ are learnable parameters, $H$ is the Heaviside function and $S$ is the sign function.

The use of $S$ and $H$ makes the use of backpropagation difficult, as their gradient is $0$ everywhere except in $0$ where it is $\infty$. However, there exist solutions to overcome this problem, and we chose the \textit{surrogate gradient} approach: the non-differentiable functions are used in the forward pass, but the derivatives of other functions, which are differentiable and similar to the non-differentiable ones, are used in the backward pass. This technique is notably used in the context of \textit{spiking neural networks} \citep{neftci2019Surrogate, eshraghian2023Training}.

\begin{figure}
    \centering
    \includegraphics{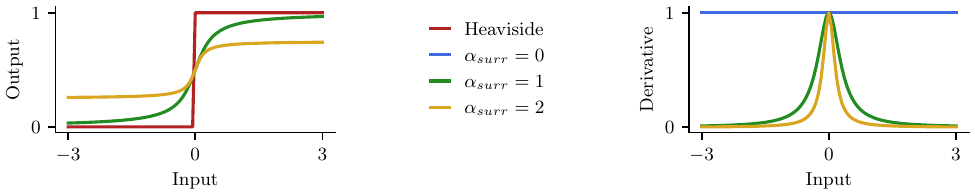}
    \caption{\textbf{Surrogate gradient used in \cell{}.} (\textbf{left}) Comparison between the Heaviside function and the function defined by \cref{eq:surr_func} used to approximate the gradient for different values of $\alpha_{surr}$. (\textbf{right}) Impact of $\alpha_{surr}$ on the surrogate gradient defined by \cref{eq:surrogate}.}
    \label{fig:surrogate}
\end{figure}

For the Heaviside function, we chose the following surrogate gradient (inspired from \citep{eshraghian2023Training})
\begin{equation}
    \frac{df}{dx}(x) = \frac{1}{1 + (\alpha_{surr} \pi x)^2} \label{eq:surrogate},
\end{equation}
where $\alpha_{surr}$ is a tunable parameter. We note that this surrogate derivative comes from the function
\begin{equation}
    f(x) = \frac{1}{\pi\alpha_{surr}} \text{atan}(\alpha_{surr} \pi x) + \frac{1}{2}. \label{eq:surr_func}
\end{equation}

\Cref{fig:surrogate} plots $f(x)$ and $\frac{df}{dx}(x)$ for different values of $\alpha_{surr}$. The derivative exhibits a localized peak centered at x = 0. As the parameter $\alpha_{surr}$ increases, the width of this peak decreases, making it more concentrated. The case $\alpha_{surr} = 0$ is special and leads to a constant derivative equal to $1$, which is also called the \textit{straight-through estimator} \citep{bengio2013Estimating}. The surrogate gradient used for $S$ is simply $2 \cdot \frac{df}{dx}(x)$, as $S$ has the same shape as $H$ with outputs being either $-1$ or $1$. \Cref{app:alpha_surr} shows how $\alpha_{surr}$ can impact the learning of BMRU networks.

\subsection{Properties of the bistable MRU}
\label{sec:properties}

\cell{} is a special recurrent cell compared to the usual ones and it has some interesting properties that are highlighted here.

\begin{figure}
    \centering
    \begin{subfigure}{.32\textwidth}
        \centering
        \includegraphics{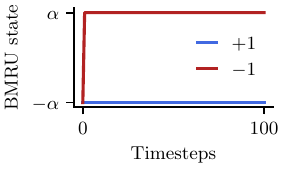}
    \end{subfigure}
    \hfill
    \begin{subfigure}{.32\textwidth}
        \centering
        \includegraphics{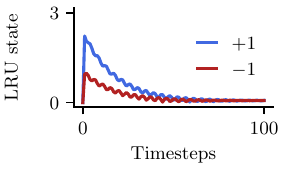}
    \end{subfigure}
    \hfill
    \begin{subfigure}{.32\textwidth}
        \centering
        \includegraphics{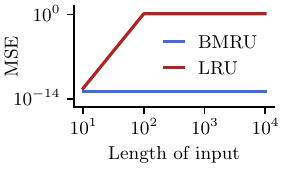}
    \end{subfigure}
    \caption{\textbf{Persistent memory of \cell{} compared to the fading memory of LRU.} Small models with one LRU or \cell{} layer of one unit have been trained on a simple benchmark whose inputs start with $\pm1$ followed by $0$'s. The goal of the models is to output the first input at the last timestep. (\textbf{left, center}) Evolution of the states of LRU and \cell{} with respect to the timesteps for the two possible inputs. As the LRU state is complex, its norm is plotted. (\textbf{right}) MSE computed with both models on the two possible inputs for different sequence lengths.}
    \label{fig:stationary_memory}
\end{figure}

\paragraph{Stationarity} \cell{} is by design stationary, meaning that repeating the same input several times will not impact state value. It makes state update independent of input duration when the input is constant. One advantage that comes from this stationarity property is the ability to generalize to longer inputs. \Cref{fig:stationary_memory} illustrates this on a simple benchmark, where the goal is to retain a binary value $\pm 1$ given at $t=0$ for some time, during which the input is $0$. This benchmark can be easily solved using either persistent memory (\cref{fig:stationary_memory}, left) or fading memory (\cref{fig:stationary_memory}, center). However, \cell{} encoding the information in stable states, it remains encoded forever, and the performance is not impacted by the input length (\cref{fig:stationary_memory}, right). LRU however encodes the information in fading memory, which does not generalize to larger input length, as memory fades over time.

\paragraph{No vanishing nor exploding gradient in the memory region} At every timestep, the state $h_t$ will be either set equal to $h_{t-1}$ or to a value independent of $h_{t-1}$. When the previous state is kept, the derivative of the new state with respect to the previous one is $1$. When the memory is updated, the derivative is $0$, and therefore the gradient will not go further back in time during the backward pass. In other words, the information is either kept intact, or overwritten. Mathematically, it writes
\begin{equation*}
    \frac{\partial h_t}{\partial h_{t-1}} = \begin{cases}
        1 & \text{ if } z_t = 0, \\
        0 & \text{ if } z_t = 1.
    \end{cases}
\end{equation*}

\begin{figure}
    \centering
    \begin{subfigure}{.32\textwidth}
        \centering
        \includegraphics{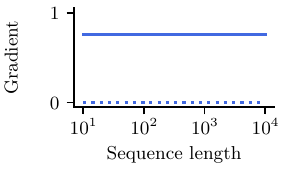}
    \end{subfigure}
    \hfill
    \begin{subfigure}{.32\textwidth}
        \centering
        \includegraphics{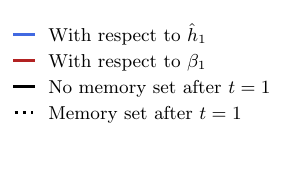}
    \end{subfigure}
    \hfill
    \begin{subfigure}{.32\textwidth}
        \centering
        \includegraphics{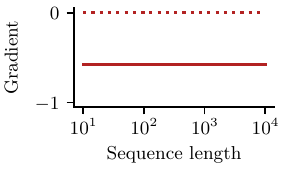}
    \end{subfigure}
    \caption{\textbf{Consistency of the gradient of \cell{} with respect to time.} Evolution of the gradient of the last state $h_T$ with respect to the first candidate $\hat{h}_1$ or $\beta_1$. Two cases are considered: either only the first timestep sets the memory, i.e. $|\hat{h}_1| > \beta_1$ and $|\hat{h}_t| < \beta_t \; \forall t > 1$, either another timestep also sets the memory.}
    \label{fig:grads}
\end{figure}

\Cref{fig:grads} shows how the gradient of the last state $h_T$ with respect to the first candidate $\hat{h}_1$ or the first $\beta_1$ evolves with respect to the sequence length. As expected, it stays constant, with two possible values: a null one if the memory has been updated at a later timestep, or a non-zero value if the memory was never updated after the first timestep. Indeed, in this case, each derivative $\frac{\partial h_t}{\partial h_{t-1}}$ is equal to 1, therefore their product is equal to $1$, ensuring a constant gradient across time.

\paragraph{Compatibility with parallel scan} \cell{} update equations can be rewritten using an associative operator, therefore allowing the use of the parallel scan, as proved in \cref{app:parallel_scan}.
A \textit{scan} is an operation that takes a binary operator $\cd$ and an ordered set of $n$ elements $[a_0, ..., a_{n-1}]$ and returns the ordered set
\begin{equation*}
    \lbr a_0, (a_0 \cd a_1), (a_0 \cd a_1 \cd a_2), \dots, (a_0 \cd a_1 \cd \dots \cd a_{n-1}) \rbr.
\end{equation*}
The goal of the parallel scan is to perform a scan with a better time complexity than by performing it sequentially, decreasing the complexity of computing the $n$ outputs from $O(n)$ to $O(\log(n))$ on a GPU.

\section{Experiments}
\label{sec:experiments}

This section aims at analyzing the performance of \cell{} on several regression and classification benchmarks, with increasing level of difficulty. The idea is to showcase the properties, strengths but also the limitations of \cell{}. Networks of LRU \citep{orvieto2023Resurrecting} were also trained to highlight differences between the SSMs and \cell{}. Each value presented in this section is computed by taking the average over 5 runs.

\subsection{Model architecture}
The architecture follows the one used in the SSM literature \citep{gu2022Efficiently,orvieto2023Resurrecting}. It consists of a linear layer that projects the input to some dimensions (called the \textit{model dimensions}, $H$), followed by recurrent blocks and fully-connected layers applied timestep-wise to extract the predictions. Each recurrent block has a batch normalization layer, recurrent cell (either \cell{} or LRU), a GLU activation and a skip connection. These recurrent cells have their own number of dimensions (\textit{state dimensions}, $N$), which can differ from the model dimensions. More specifically, they receive inputs in $H$ dimensions, update their $N$ dimensional state and generate an output with $H$ dimensions. The recurrent cells can also receive positional embeddings that encode the current timestep into some vector of arbitrary dimensions. Unless it is specified otherwise, we consider the prediction of the model to be its output at last timestep, as our main goal is to evaluate the ability of the models to retain information. The parameters of each experiment are given in \cref{app:training_parameters}.

\subsection{Copy-first-input}
\label{sec:cfi_exp}

The first benchmark is purely synthetic and consists of remembering a real value for some duration while being perturbed by noise. In practice, models receive a two-dimensional input $x_t = (r_t, f_t)$ at each timestep. The first dimension is a real value $r_t$ independently and identically distributed from N(0,1) at all timesteps. The second one is a flag $f_t$ indicating if $r_t$ has to be retained ($f_t = 1$) or not ($f_t = 0$). This flag is only set at the first timestep. Models are trained by computing the MSE between the first input and their last output. In this experiment, we used networks of 2 recurrent blocks, with $H = N = 256$ and without any positional encoding. All the models are trained on $60000$ samples, $10\%$ of which are used for validation. The test set also consists of $60000$ samples.

\begin{figure}[tbp]
    \begin{minipage}[t]{0.32\textwidth}
        \centering
        \includegraphics{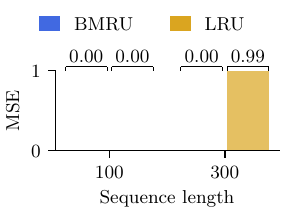}
        \caption{\textbf{Results on the copy-first input benchmark.} Test MSEs obtained by \cell{} and LRU models on the copy-first-input benchmark for two sequence lengths: $100$ and $300$.}
        \label{fig:cfi}
    \end{minipage}
    \hfill
    \begin{minipage}[t]{0.64\textwidth}
        \centering
        \includegraphics{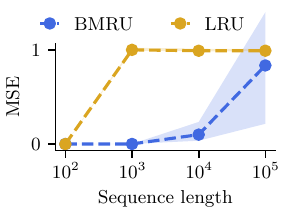}
        \hfill
        \includegraphics{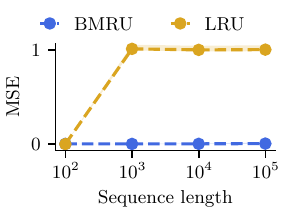}
        \caption{\textbf{Generalization capabilities of \cell{} and LRU with respect to the sequence length.} MSEs obtained by the \cell{} and LRU models trained on the copy-first-input with a sequence length of 100 when they are evaluated on longer sequences with two different levels of noise. (\textbf{left}) The noise is sampled from $\mathcal{N}(0, 1)$. (\textbf{right}) The noise is sampled from $\mathcal{N}(0, 0.1)$.}
        \label{fig:cfi_generalization}
    \end{minipage}
\end{figure}

\Cref{fig:cfi} shows the test MSE obtained by the networks on two versions of the benchmark, one where information has to be retained for $100$ timesteps, the other for $300$ timesteps. For the small duration, both models learn. However, for the longer duration, only \cell{} manages to learn the task. Although increasing network depth would possibly make LRU learn, our goal here is not to optimize the hyperparameters to get the best loss possible, but rather to highlight the ability of \cell{} to retain information for a long duration even with a shallow network.

This benchmark also allows to highlight generalization property of \cell{} to longer sequences. One can evaluate this generalization by taking the models trained on $100$ timesteps and testing them on longer sequences. To do that, we created small test sets whose sequences lengths are a power of $10$, starting from $10^2$ to $10^5$. Each test set is composed of $6000$ samples. \Cref{fig:cfi_generalization} shows the evolution of the MSEs obtained by both cells on these small test sets with respect to the sequences length. LRU does not generalize to larger sequences well, as it encodes the information in fading memory. \cell{} is much more resilient as it encodes the information in persistent memory, and the performance only decreases if the stored values are overwritten by high amplitude noise (center graph). If the noise is sufficiently low, the performance is unaffected by the sequence length (right graph).

\subsection{Permuted sequential MNIST}
\label{sec:seqmnist_exp}

The MNIST dataset is one of the most well-known datasets \citep{lecun1998Gradientbased}. In this experiment, we use a variant of MNIST, called the Sequential MNIST \citep{le2015Simple}, where the pixels are fed to the models one by one. While being known as an \textit{easy} benchmark, the Sequential MNIST is still relevant when testing new architectures. To increase the difficulty, the pixels are shuffled before being given to the models. Compared to the previous benchmark, models must combine information received at different timesteps in order to predict the correct label. To also assess the memorization capabilities of the models, we add, in some experiments, black pixels at the end of the sequences. This forces the models to not only combine temporal information but also to retain it for a long period. In practice, $1216$ black pixels are added, bringing the sequence lengths to $2000$. Furthermore, we add positional encodings to the inputs of \cell{}, as we observed that it can greatly improve its performance. On the other hand, we observed that adding these encodings to LRU actually impedes its performance, therefore these are only given to \cell{} in these experiments.

\begin{figure}[tbp]
    \centering
    \begin{subfigure}[t]{.49\textwidth}
        \centering
        \includegraphics{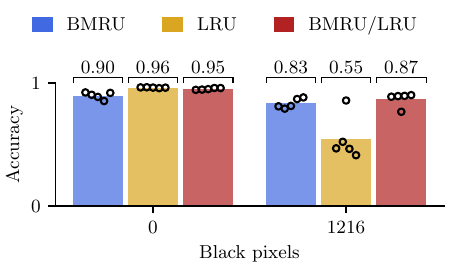}
        \caption{2 rec. blocks}
        \label{fig:seqmnist_nr2}
    \end{subfigure}
    \begin{subfigure}[t]{.49\textwidth}
        \centering
        \includegraphics{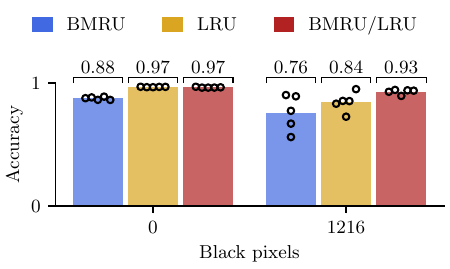}
        \caption{3 rec. blocks}
        \label{fig:seqmnist_nr3}
    \end{subfigure}
    \caption{\textbf{Results on the permuted sequential MNIST.} Accuracies obtained by \cell{}, LRU and \cell{}/LRU models on the permuted sequential MNIST, with or without black pixels added at the end of the sequences. Models with 2 and 3 recurrent block have been tested.}
    \label{fig:seqmnist}
\end{figure}

\Cref{fig:seqmnist} shows the accuracies obtained by \cell{} (blue bars) and LRU models (yellow bars), with or without black pixels, and for two network depths. All recurrent blocks contain $256$ neurons, as in the previous experiment. The performance of LRU is better than \cell{} on the version without black pixels, but \cell{} better maintains its performance when black pixels are added, especially for the more shallow model. LRU indeed requires more depth to handle longer dependencies, whereas \cell{} only needs two layers to handle these long dependencies. On the other hand, this experiment also shows that LRU is much better at combining the information from the different timesteps. This motivates our next experiment: as LRU is better at combining the information using fading memory, and \cell{} at retaining it using persistent memory, their combination should get the best of both worlds, while still being parallelizable. The results obtained using this combination are shown in \cref{fig:seqmnist} (red bars). Note that for a fair comparison, the state dimension of each cell has been divided by $2$, in order to maintain the number of neurons in each recurrent block at $256$. We can see that the combination has the same performance than LRU alone when there are no black pixels, but its performance does not decrease when black pixels are added. It even stays higher than \cell{} alone. This highlights the potential of combining the fading memory and persistent memory properties of these two types of parallelizable RNNs. For more studies on this hybrid layer, \cref{app:hybrid_ratio} evaluates the impact of using different ratios of BMRU/LRU neurons. Moreover, \cref{app:alpha_bmru} studies how making the parameter $\alpha$ input-dependent influences performance.

\subsection{Pathfinder}
\label{sec:pf_exp}

The pathfinder benchmark is part of the \textit{long-range arena} group of benchmarks \citep{tay2020Long}. It consists of 32x32 black and white images where lines are drawn randomly, as well as two dots. The goal of the benchmark is to predict if the two dots are connected by a line or not. Images are fed pixel by pixel, which leads to sequences of $1024$ timesteps. The goal of this experiment is to test the capabilities of shallow \cell{} networks, i.e. max 3 recurrent blocks, to solve such a difficult benchmark without trying to achieve state-of-the-art performance through parameter tuning. We chose a similar model approach as in the S4 and LRU papers \citep{gu2022Efficiently,orvieto2023Resurrecting}: the predictions of the models are the means of their outputs (instead of their last timestep), and the recurrent layers are bidirectional.

\begin{figure}[tbp]
    \centering
    \begin{subfigure}{.32\textwidth}
        \centering
        \includegraphics{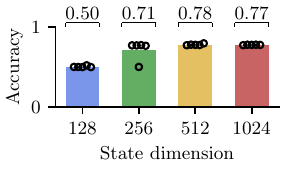}
        \caption{1 rec. block}
        \label{fig:pathfinder_sbc_nr1}
    \end{subfigure}
    \begin{subfigure}{.32\textwidth}
        \centering
        \includegraphics{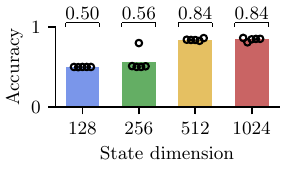}
        \caption{2 rec. blocks}
        \label{fig:pathfinder_sbc_nr2}
    \end{subfigure}
    \begin{subfigure}{.32\textwidth}
        \centering
        \includegraphics{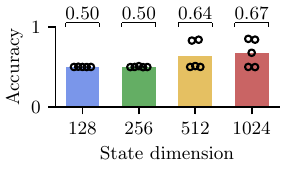}
        \caption{3 rec. blocks}
        \label{fig:pathfinder_sbc_nr3}
    \end{subfigure}
    \caption{\textbf{Results on the pathfinder benchmark.} Each plot shows the accuracies obtained by \cell{} models on the pathfinder benchmark, for different state dimensions (x-axes) and network depths (1, 2 and 3 recurrent blocks from left to right, respectively).}
    \label{fig:pathfinder}
\end{figure}

\Cref{fig:pathfinder} shows the accuracies obtained by \cell{} on this benchmark. Note that as there are only two classes, so an accuracy of $50\%$ corresponds to random guess. First, the results show that the \cell{} is able to learn on the Pathfinder benchmark, even with only one recurrent layer. Second, these experiments show that the depth of the network is not that important, but its width, i.e. the state dimension, has a bigger impact on performance. This differs from SSMs, for which it is known that their ability to handle longer time dependencies grows with the network depth. The drop of performance when more layers are added can be explained by the \textit{discretization} implied by \cell{} and the simplicity of layer interconnections. Indeed, as each neuron of \cell{} can only output two values ($\pm \alpha$), each \cell{} layer adds more discretization, and makes the learning more difficult.

\section{Discussion}
\label{sec:discussion}

The concept of MRU and \cell{} are at the junction of several topics that are discussed in this section and compared to relevant works.

\paragraph{Multistable RNNs} Adding multistability in RNNs is not a very explored topic from the point of view of machine learning. Some works have been done to better understand multistability in RNNs from a dynamical system point of view \citep{cheng2006Multistability,krishnamurthy2022Theory}, even some that includes \textit{thresholding} functions \citep{edwards2000Analysis,tournoy2023Step}. However, these works do not include any machine learning experiments. Few works have highlighted how this property can improve the memorization capability of RNNs, either by building new types of RNNs \citep{vecoven2021bioinspired} or by enforcing multistability in existing RNNs \citep{lambrechts2023Warming}. This work attempts to highlight the interesting properties of multistable RNNs for sequence learning and its complementarity with the more classical fading memory.

\paragraph{Parallelizable RNNs} Parallelization is a \textit{sine qua non} condition nowadays, and is one the features of Transformers that put them in the front of the scene. This motivates also the research in SSMs. Since the creation of SSMs \citep{gu2021Combining, gu2022Efficiently}, numerous papers have added their contributions and improvements, which finally led to Mamba \citep{gu2024Mamba}, the first SSM architecture that was used in large language models.
Some approaches augment SSMs to improve their capacity to handle temporal dependencies. For instance, \citet{sun2025Parallel} adds a second smaller SSM inside the usual one. This new SSM generates a gate at every timestep, which is then used to combine the $n$ previous states into one vector. This approach reminds of the attention mechanism, and while it helps RNNs at remembering past information, it does not solve the fading memory problem. However, it provides a nice way to combine information from nearby timesteps and might complement BMRU to alleviate its weaknesses.
Other works tried to make RNN parallelizable, but most of them end up removing the nonlinearities \citep{martin2018Parallelizing,beck2024xLSTM,feng2024Were}.
There exist works that developed parallelizable RNNs, but not without concession: They had for example to remove the time-dependency in the RNN update \citep{jiang2021Recurrent} or to limit the depth of the network \citep{zini2021Optimized}. Recent and promising approaches reformulate the RNNs equations as a system and use iterative methods to solve it \citep{lim2023Parallelizing,gonzalez2024Scalable,danieli2025ParaRNN}. In this work, we show that multistable RNNs can be parallelizable over the sequence length if we consider the case of convergent RNNs, i.e. RNNs that reach full convergence between two timesteps.

\paragraph{Surrogate gradients} The usage of non-differentiable functions in neural networks does not happen often, therefore surrogate gradients are not that useful in classic deep learning. However, in some fields with restrictions, it allows to benefit from all the advantages of back-propagation while using special functions. For instance, in the topic of \textit{spiking neural networks}, where outputs must be 0's and 1's, surrogate gradients have allowed the networks to be trained like any classical networks \citep{eshraghian2023Training,lee2016Training,nieves2021Sparse,neftci2019Surrogate}. However, the impact of using an approximation during backward pass is difficult to measure, and therefore makes the usage of surrogate gradients delicate. Also referred to as \textit{pseudo-gradient}, this idea is not really new: \citet{bengio2013Estimating} introduces the \textit{straight-through estimator} uses a constant pseudo-gradient, i.e. the derivative of a linear activation, while \citet{zeng1993Learning} and \citet{goodman1994learning} use the derivative of a sigmoid to train MLPs and RNNs with threshold units.

\paragraph{Steady-states and equilibria} While \cell{} does not really rely on the computation of steady-states, the notion of MRU is built around it and the concept of implicit function. This reminds of \textit{Deep Equilibrium Models} (DEQs) introduced by \citet{bai2019Deep} where layers are formulated as an implicit function and outputs are the steady-states of this implicit function. They use a solver to estimate this steady-state starting from some initial guess. However, this initial guess does not depend on any past information, and the implicit function are typically monostable, therefore these layers do not implement any memory. A MRU can be seen as a variant of DEQs where the implicit function is multistable and the initial guess is chosen to be the previous steady-state.

\paragraph{Binary gates} BMRU emulates the bistability property through the binary gate $z$. Other works also rely on binary gates to improve the memorization of RNNs. For instance, \citet{yin2026Efficient} generates a sequence of binary gates with a rhythmic module, allowing to not update all the components of the hidden state at every timestep. An other example is the work of \citet{shan2023DRRNets} that introduces a binary gate and a specific training algorithm to learn optimal recurrent routes.

\paragraph{Improving MRUs and \cell{}} While BMRU already achieved appreciable performance on several benchmarks, some concerns may be raised about its capacity to address more complex tasks. Indeed, we already mention its difficulty to scale to deeper networks, as well as the computational limitation imposed by the quantization and finiteness of the state space of BMRU. However, more recent works based on MRUs and BMRU have increased the learning capabilities of these approaches by creating new variants of BMRU or by improving the model architecture. These new versions allows deeper networks of MRUs to be trained and achieve excellent performance on several benchmarks including language modeling tasks. Notably, \citet{fyon2026HardwareSoftware} shows that a 6-layers BMRU network performs well on a language modeling task, and also that BMRU can be adapted to run on specific hardware without compromising performance. Furthermore, \citet{brandoit2026Improving} introduces a variant of BMRU, the cumulative MRU (CMRU), that relies on an infinite number of stable states rather than only two and achieve comparable performance to SSMs. This work also shows that the quantization of states does not impede performance. Finally, \citet{bakija2026Importance} extends the idea of generalization briefly mentioned in \cref{sec:properties} in the context of reinforcement learning and shows how BMRU allows agents to generalize to much longer environments without any retraining. It also highlights the advantage of the hybrid model in tasks that both require integrating information distributed at different timesteps and remembering information over long periods. This body of work demonstrates the great potential of MRUs in various contexts, including complex ML benchmarks, embedded AI or reinforcement learning.

\paragraph{Future works} In the experiments section, we highlighted three characteristics of \cell{}: its property to extend its memory to much longer durations, the gain of performance that can be obtained when combining it with a SSM, and its ability to learn long-term dependencies in difficult benchmarks with shallow networks. All of these deserve to be explored.
For instance, we know that SSMs are better in deep networks, which is not the case of \cell{}, therefore making mixed models where recurrent blocks are made of more SSM layers than \cell{} layers could lead to interesting results.
Also, this degradation of performance when more \cell{} layers are added could be explored, as improvements in the cell equations, the initialization or the training algorithm \citep{zhang2026Aligning} could be made to improve learning in deeper models. As mentioned previously, the works of \citet{brandoit2026Improving,fyon2026HardwareSoftware} already improve the scaling of MRUs to deeper networks.
Furthermore, \cell{} equations have three interesting properties: the quantization of the state, the shape of the update decision, which is a comparison between $\cand_t$ and $\beta_t$, and finally the possibility for the update gate $z_t$ to be null. The first one is necessary to have discrete stable states, the second one to approximate the hysteresis bifurcation and the last one to have persistent memory (as $z_t = 0$ implies $h_t = h_{t-1}$). However, these can be used independently of the others. The impact of each definitively deserves to be explored, especially the third one which allows for persistent memory, as we are not aware of another RNN architecture that allows for this property.
In addition to that, \cell{} update rule is restrictive: either we keep the past state, or we totally forget it. However, nothing prevents from combining the past state and the input, as long as the dependency with respect to the past state stays linear. The CMRU mentioned in the previous section is an example of this.
Also, we note that for introducing the MRU, we put forward a concept of RNNs with internal clocks, the MRU being a specific implementation of such cells when the internal clock iterates an infinite number of times. In this respect, we believe it may also be potentially interesting to further exploit this concept when the clock iterates a finite number of times. Furthermore, the usage of surrogate gradients is practically inexistent in classic deep learning, as all computations are differentiable. While it seems that using surrogate gradients may impede the performance of neural networks by introducing some mismatches between forward and backward passes, we wonder if using non-differentiable functions in classical neural networks may increase the performance and robustness of these networks. Indeed, step functions like the Heaviside one are more resilient to small changes in their output, but they probably make the training more difficult. Finally, \cell{} was only tested on classification and regression tasks, but one of its advantages shared with SSMs is its efficient sequence generation capacity. It would be interesting to try it on generation tasks like text generation benchmarks for instance, or in reinforcement learning tasks as done by \citet{bakija2026Importance}.

\section{Conclusion}

In this paper, we introduce the concept of \textit{memory recurrent units} (MRU): a new class of RNNs that do not exhibit any transient dynamics but that creates persistent memory through multistability. We also present a concrete implementation of a MRU, the \textit{\cellname{}} (\cell{}), derived from the hysteresis bifurcation. The equations of \cell{} are closed to the usual gated RNN equations, and are compatible with the parallel scan algorithm.

We observe that \cell{} can achieve good results in different benchmarks, all requiring learning long-term dependencies. Moreover, combining \cell{} with a SSM leads to a parallelizable recurrent model that has both linear transient dynamics and a multistable behavior, which allows to efficiently encode temporal information for very long durations. Indeed, the linear dynamics can encode complex information but this memory will unavoidably fade over time, while multistability encode more qualitative information in a never-fading memory.

Finally, \cell{} has shown to be efficient with shallow networks, while SSMs typically requires more layers to learn long-term dependencies.

To conclude, MRUs are a new concept that deserve to be explored and especially since we showed it could lead to a well working implementation, the \cell{}, that showed interesting properties and performance. This paves the way for new experiments and improvements for future designs.

\section*{Acknowledgments}
This work has been the subject of patent applications (Numbers: EP26151077 and EP26175248.9).
Florent De Geeter gratefully acknowledges the financial support of the Walloon Region for Grant No. 2010235 – ARIAC by DW4AI.
Gaspard Lambrechts is a postdoctoral researcher of the \emph{Fund for Scientific Research} (FNRS) from the \emph{Wallonia-Brussels Federation} in Belgium.
The present research benefited from computational resources made available on Lucia, the Tier-1 supercomputer of the Walloon Region, infrastructure funded by the Walloon Region under the grant agreement n°1910247. This work was supported by the Belgian Government through the Federal Public Service Policy and Support.

\bibliography{bib/bib.bib}

\newpage
\appendix

\section{Hysteresis bifurcation in the bistable recurrent cell}
\begin{figure}[tbp]
    \centering
    \includegraphics{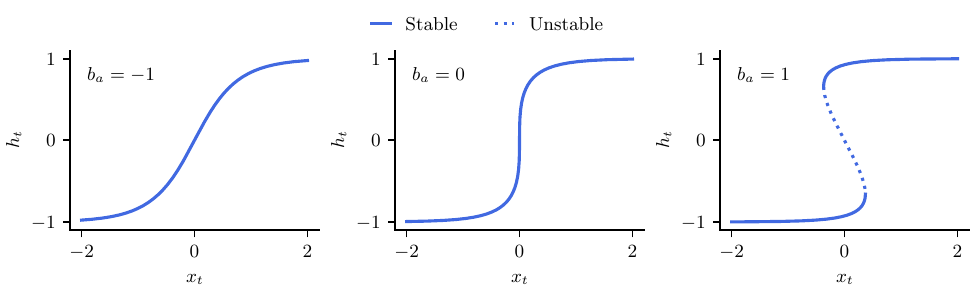}
    \caption{\textbf{Hysteresis bifurcation in the bistable recurrent cell.} This figure shows the solutions to the implicit function defined by \cref{eq:BRC_static} as well as their stability for three values of $b_a$.}
    \label{fig:BRC_hyst}
\end{figure}

\label[appendix]{app:Hyst_BRC}
The bistable recurrent cell \citep{vecoven2021bioinspired} is described the set of equations
\begin{subequations}
    \begin{align}
        c_t & = \sigma(U_c x_t + w_c \odot h_{t-1} + b_c),  \label{eq:BRCa}                            \\
        a_t & = 1 + \tanh{U_a x_t + w_a \odot h_{t-1} + b_a},  \label{eq:BRCb}                         \\
        h_t & = c_t \odot h_{t-1} + (1- c_t) \odot \tanh{U_h x_t + a_t h_{t-1} + b_h}, \label{eq:BRCc}
    \end{align}
\end{subequations}
where $c_t$ is the update gate, $a_t$ the feedback gate, and $U_c, w_c, b_c, U_a, w_a, b_a, U_h, b_h$ are learnable parameters. At steady-state, we have $h_t = h_{t-1}$, and \cref{eq:BRCc} writes
\begin{equation*}
    (1- c_t) \odot h_t = (1- c_t) \odot \tanh{U_h x_t + a_t h_{t} + b_h},
\end{equation*}
which, by replacing $a_t$ by \cref{eq:BRCb} at steady-state and dividing each term by $(1- c_t)$ ($c_t$ being the output of a sigmoid function, its value is strictly smaller than 1), leads to
\begin{equation}
    \tanh{U_h x_t + (1 + \tanh{U_a x_t + w_a \odot h_{t} + b_a}) h_{t} + b_h} - h_t =0. \label{eq:BRC_static}
\end{equation}
This steady-state function is a static, implicit function that relates the values of the input $x_t$ to the values of the state $h_t$ at convergence. We can show that this function corresponds to a hysteresis singularity with $x_t$ as the bifurcation parameter and $b_a$ as the unfolding parameter. \Cref{fig:BRC_hyst} illustrates the unfolding of the hysteresis bifurcation for $b_a = -1$ (monostable), $b_a = 0$ (singular) and $b_a = 1$ (bistable) (other parameter values are $U_h=1$, $U_a=w_a=b_h=0$). Modifications of the parameters $U_h\neq0, U_a, w_a, b_h$ lead to translations and deformations of the bifurcation plot without alteration of the existence of a hysteresis singularity.

\section{Compatibility of \cell{} with the parallel scan algorithm}
\label[appendix]{app:parallel_scan}

The \textit{scan} operation, also called the \textit{prefix sum} operation, computes from a binary operator $\oplus$ and an array of $n$ elements $[a_0, a_1, \dots, a_{n-1}]$ the array
\begin{equation*}
    [a_0, a_0 \oplus a_1, \dots, a_0 \oplus a_1 \oplus ... \oplus a_{n-1}].
\end{equation*}

A naive implementation of the scan consists in a simple loop over the input array that accumulates the results and stores them in the output array. This implementation has a time complexity of $O(n)$. However, it is possible to perform the scan with a better time complexity when the operator $\oplus$ is associative and when multiple processors are available.

The parallel scan algorithm introduced by \citet{blelloch1990Prefix} is an algorithm that performs the \textit{scan} operation on a array of $n$ elements with a time complexity of $O(\frac{n}{p} + \log(p))$, where $p$ is the number of processors. Assuming this algorithm is run on a GPU, we can consider $p = n$, which therefore gives a time complexity of $O(\log(n))$. Notably, this algorithm is used with the SSMs during their training \citep{smith2022Simplified}.

This section proves that \cell{} equations are compatible with the parallel scan algorithm.

\begin{theorem}
    The parallel scan can be used to perform the scan on an array with a binary operator $\oplus$ if and only if $\oplus$ is associative.
\end{theorem}

\begin{proof}
    The proof is given in \citep{blelloch1990Prefix}.
\end{proof}

Furthermore, \citet{blelloch1990Prefix} showed that there exists a binary associative operator that computes the ordered set of states from a first-order linear recurrence.

\begin{theorem}
    \label{th:lin_rec}
    Assume a first-order linear recurrence of the following form:
    \begin{equation*}
        h_t = \begin{cases}
            b_0                     & \text{ if } t = 0,     \\
            a_t \odot h_{t-1} + b_t & \text{ if } 0 < t < T,
        \end{cases}
    \end{equation*}
    where $a_t$ and $b_t$ $\forall t \in [0, T]$ are independent of $h_{t'}$ $\forall t' \in [0, T]$. Consider the set of pairs $c_t = [a_t, b_t]$ and the binary operator $\cd$ defined as follows:
    \begin{equation*}
        c_t \cd c_{t'} \equiv [c_{t, a} \odot c_{{t'}, a}, c_{{t'}, a} \odot c_{t, b} + c_{{t'}, a}]
    \end{equation*}
    where $c_{t, a}$ and $c_{t, b}$ are the first and second elements of $c_t$.
    Then,
    \begin{enumerate}
        \item The operator $\cd$ is associative,
        \item Performing the scan with the operator $\cd$ on the array $[c_0, \dots, c_T]$ creates the array $[s_0, \dots, s_T]$ where $s_t = [y_t, h_t]$ and $y_t$ is defined as:
              \begin{equation*}
                  y_t = \begin{cases}
                      a_0               & \text{ if } t = 0,     \\
                      a_t \odot y_{t-1} & \text{ if } 0 < t < T.
                  \end{cases}
              \end{equation*}
    \end{enumerate}

    It results that the parallel scan can be used to solve this first-order linear recurrence as $\cd$ is associative (point 1.), and the solutions $h_t$ will be the second values of the generated pairs $s_t$ (point 2.).
\end{theorem}

\begin{proof}
    The proof is given in \citep{blelloch1990Prefix}, section 4.1.
\end{proof}

To prove that \cell{} is compatible with the parallel scan, we can therefore show that \cref{eq:cell_cand,eq:cell_beta,eq:cell_z,eq:cell_h} can be rewritten as a first-order linear recurrence.

\begin{theorem}
    Assume a sequence of inputs $[x_1, \dots, x_T]$ and fixed parameters $W_x, W_{\beta}, b_x, b_{\beta}$ and $\alpha$. \Cref{eq:cell_cand,eq:cell_beta,eq:cell_z,eq:cell_h} describe a first-order linear recurrence, and are therefore compatible with the parallel scan.
\end{theorem}

\begin{proof}
    First, let us separate the equations that are independent of the previous state $h_{t-1}$ from the ones that are not. \Cref{eq:cell_cand,eq:cell_beta,eq:cell_z} do not depend on $h_{t-1}$. Therefore $\cand_t$, $\beta_t$ and $z_t$ can be computed in parallel for all timesteps.

    The last equation, \cref{eq:cell_h}, is the only one that has to be analyzed. As a reminder, here it is:
    \begin{equation*}
        h_t = z_t \odot S(\cand_t) \odot \alpha + (1 - z_t) \odot h_{t-1}.
    \end{equation*}

    The left term is independent of $h_{t-1}$, hence let us define the set of $b_t$'s as follows:
    \begin{equation*}
        b_t \equiv z_t \odot S(\cand_t) \odot \alpha.
    \end{equation*}

    Also, the factor that multiplies $h_{t-1}$ is independent of it. Let us define the set of $a_t$'s as follows:
    \begin{equation*}
        a_t \equiv 1 - z_t.
    \end{equation*}

    \Cref{eq:cell_h} can therefore be rewritten using $a_t$ and $b_t$:
    \begin{subequations}
        \begin{align}
            a_t & = 1 - z_t, \label{eq:scan_a}                           \\
            b_t & = z_t \odot S(\cand_t) \odot \alpha, \label{eq:scan_b} \\
            h_t & = \begin{cases}
                        h_0                     & \text{ if } t = 0,     \\
                        a_t \odot h_{t-1} + b_t & \text{ if } 0 < t < T.
                    \end{cases} \label{eq:scan_h}
        \end{align}
    \end{subequations}

    Once again, \cref{eq:scan_a,eq:scan_b} are independent of $h_{t-1}$ and can be computed in parallel for all timesteps. Finally, \cref{eq:scan_h} is a first-order linear recurrence and by using \cref{th:lin_rec}, parallel scan can be used to solve it.
\end{proof}

\section{Training parameters}
\label[appendix]{app:training_parameters}

This section gives all the training parameters used in the experiments of \cref{sec:experiments}: \Cref{tab:cfi_params,tab:seqmnist_params,tab:pf_params} respectively give the parameters of \cref{sec:cfi_exp,sec:seqmnist_exp,sec:pf_exp}.

\begin{table}[htbp]
    \centering
    \begin{tabular}{|l|l|}
        \hline
        \textbf{Parameter}                                                   & \textbf{Value(s)}                                                                                                            \\
        \hline
        Number of samples in dataset                                         & $60000$                                                                                                                      \\
        Sequence length                                                      & $100$ - $300$                                                                                                                \\
        Train / valid ratio                                                  & $90\%$ / $10\%$                                                                                                              \\
        Epochs                                                               & 100                                                                                                                          \\
        Learning rate                                                        & \makecell[tl]{Cosine annealing:               \\$10^{-4} \to 10^{-3}$ during $10$ first epochs,\\then $10^{-3} \to 10^{-5}$} \\
        Weight decay                                                         & \makecell[tl]{$0.0001$ for \cell{} parameters \\ $0.05$ for other parameters}                                                \\
        \hline
        Number of recurrent blocks                                           & $2$                                                                                                                          \\
        Model dim                                                            & $256$                                                                                                                        \\
        State dim                                                            & $256$                                                                                                                        \\
        Positional encoding                                                  & No                                                                                                                           \\
        Activation between blocks                                            & GLU                                                                                                                          \\
        Bidirectional                                                        & No                                                                                                                           \\
        \makecell[tl]{Number of fully connected layers \\(after last block)} & $2$                                                                                                                          \\
        Pooling for prediction                                               & Last timestep                                                                                                                \\
        (\cell{}) $\alpha_{surr}$                                            & $1$                                                                                                                          \\
        (LRU) $r_{min}$, $r_{max}$ and $\theta_{max}$                        & $0.0$, $0.99$ and $2\pi$                                                                                                     \\
        \hline
    \end{tabular}
    \caption{Training parameters used in \cref{sec:cfi_exp}.}
    \label{tab:cfi_params}
\end{table}

\begin{table}[htbp]
    \centering
    \begin{tabular}{|l|l|}
        \hline
        \textbf{Parameter}                                                   & \textbf{Value(s)}                                                                                                            \\
        \hline
        Number of black pixels                                               & $0$ - $1216$                                                                                                                 \\
        Pixel normalization                                                  & $p / 255 - 0.5$, with $p \in [0, 255]$                                                                                       \\
        Train / valid ratio                                                  & $90\%$ / $10\%$                                                                                                              \\
        Epochs                                                               & 100                                                                                                                          \\
        Learning rate                                                        & \makecell[tl]{Cosine annealing:               \\$10^{-4} \to 10^{-3}$ during $10$ first epochs,\\then $10^{-3} \to 10^{-5}$} \\
        Weight decay                                                         & \makecell[tl]{$0.0001$ for \cell{} parameters \\ $0.05$ for other parameters}                                                \\
        \hline
        Number of recurrent blocks                                           & $2$ - $3$                                                                                                                    \\
        Model dim                                                            & $256$                                                                                                                        \\
        State dim                                                            & $256$                                                                                                                        \\
        Positional encoding                                                  & $16$ dim, only for \cell{}                                                                                                   \\
        Activation between blocks                                            & GLU                                                                                                                          \\
        Bidirectional                                                        & No                                                                                                                           \\
        \makecell[tl]{Number of fully connected layers \\(after last block)} & $2$                                                                                                                          \\
        Pooling for prediction                                               & Last timestep                                                                                                                \\
        (\cell{}) $\alpha_{surr}$                                            & $1$                                                                                                                          \\
        (LRU) $r_{min}$, $r_{max}$ and $\theta_{max}$                        & $0.0$, $0.99$ and $2\pi$                                                                                                     \\
        \hline
    \end{tabular}
    \caption{Training parameters used in \cref{sec:seqmnist_exp}.}
    \label{tab:seqmnist_params}
\end{table}

\begin{table}[htbp]
    \centering
    \begin{tabular}{|l|l|}
        \hline
        \textbf{Parameter}                                                   & \textbf{Value(s)}                                                                                                                       \\
        \hline
        Pixel normalization                                                  & \makecell[tl]{$(p - \mu)/\sigma$, with $p \in [0, 255]$, \\$\mu = 10.94$ and $\sigma = 38.51$}                                          \\
        Train / valid ratio                                                  & $90\%$ / $10\%$                                                                                                                         \\
        Epochs                                                               & 100                                                                                                                                     \\
        Learning rate                                                        & \makecell[tl]{Cosine annealing:                          \\$10^{-4} \to 10^{-3}$ during $10$ first epochs,\\then $10^{-3} \to 10^{-5}$} \\
        Weight decay                                                         & \makecell[tl]{$0.0001$ for \cell{} parameters            \\ $0.05$ for other parameters}                                                \\
        \hline
        Number of recurrent blocks                                           & $1$ - $2$ - $3$                                                                                                                         \\
        Model dim                                                            & $256$                                                                                                                                   \\
        State dim                                                            & $128$ - $256$ - $512$ - $1024$                                                                                                          \\
        Positional encoding                                                  & $16$ dim                                                                                                                                \\
        Activation between blocks                                            & GLU                                                                                                                                     \\
        Bidirectional                                                        & Yes                                                                                                                                     \\
        \makecell[tl]{Number of fully connected layers \\(after last block)} & $2$                                                                                                                                     \\
        Pooling for prediction                                               & Mean of all timesteps                                                                                                                   \\
        (\cell{}) $\alpha_{surr}$                                            & $1$                                                                                                                                     \\
        \hline
    \end{tabular}
    \caption{Training parameters used in \cref{sec:pf_exp}.}
    \label{tab:pf_params}
\end{table}

\section{\texorpdfstring{Impact of $\alpha_{surr}$ on the performance}{Impact of ⍺_surr on the performance}}
\label[appendix]{app:alpha_surr}

\begin{figure}[tbp]
    \centering
    \begin{subfigure}[t]{.49\textwidth}
        \centering
        \includegraphics{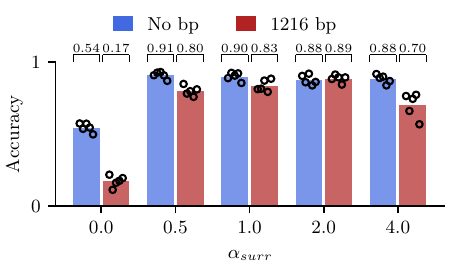}
        \caption{\cell{} alone}
        \label{fig:seqmnist_sbc_surr}
    \end{subfigure}
    \begin{subfigure}[t]{.49\textwidth}
        \centering
        \includegraphics{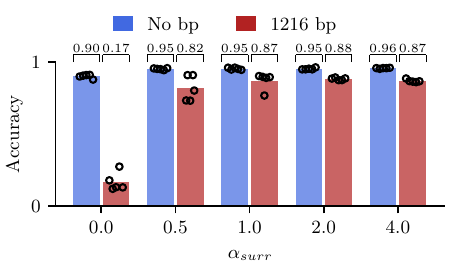}
        \caption{Hybrid model}
        \label{fig:seqmnist_both_surr}
    \end{subfigure}
    \caption{\textbf{Impact of $\alpha_{surr}$ on the performance.} Accuracies obtained by \cell{} and \cell{}/LRU models on the permuted sequential MNIST, with or without black pixels added at the end of the sequences, and for different values of $\alpha_{surr}$. All models have two recurrent layers with a size of $256$.}
    \label{fig:seqmnist_surr}
\end{figure}

In \cref{sec:a_learnable_bmru} we introduced the concept of \textit{surrogate gradient} as a solution to the impossibility to train \cell{} because of its use of functions with ill-defined derivatives, i.e. the Heaviside and the sign functions. The surrogate gradient used in the experiments is defined by \cref{eq:surrogate} and consist of a peak centered at $x=0$ with a maximal value of $1$. The tunable parameter $\alpha_{surr}$ dictates the width of this peak, as shown in \cref{fig:surrogate}. The case $\alpha_{surr} = 0$ is a special case as it leads to a constant derivative of $1$. We used $\alpha_{surr} = 1$ in the experiments of \cref{sec:experiments}. However, it may have an impact on the performance reached by the models. This appendix therefore evaluate this impact by training different models (with only \cell{} or hybrid) on the permuted sequential MNIST (with and without black pixels) for different values of $\alpha_{surr}$.

\Cref{fig:seqmnist_surr} shows the mean accuracies obtained by these models and emphases how adding black pixels can degrade more or less degrade the performance depending on $\alpha_{surr}$. A null value leads to poor performance, which suggests the importance of weighting the gradients depending on how close the inputs are from the step. The other values all lead to models that learn well, but the extreme values, i.e. $0.5$ and $4.0$ tends to leads to worse accuracies. The two last values, $1.0$ and $2.0$, give similar results.

\section{\texorpdfstring{Impact of the \cell{}/LRU ratio on the performance}{Impact of the BMRU/LRU ratio on the performance}}
\label[appendix]{app:hybrid_ratio}

\begin{figure}[tbp]
    \centering
    \includegraphics{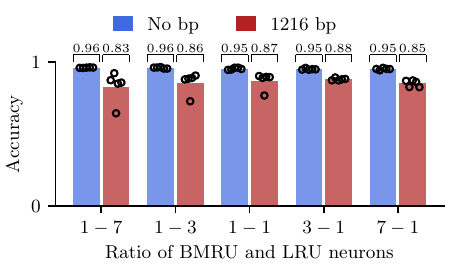}
    \caption{\textbf{Impact of the \cell{}/LRU ratio on the performance.} Accuracies obtained by hybrid models on the permuted sequential MNIST, with or without black pixels added at the end of the sequences, and for different ratios of \cell{} and LRU neurons in each recurrent layer. $X - Y$ must be read as $X$ \cell{} neurons for $Y$ LRU neurons. All models have two recurrent layers with a size of $256$.}
    \label{fig:seqmnist_hybrid_ratio}
\end{figure}

In \cref{sec:seqmnist_exp} we combined \cell{} with LRU to create hybrid recurrent layers, with the goal of bringing together their strengths and mitigating their weaknesses. The combination indeed helped improving the performance of models. However, we only tested hybrid layer including the same proportion of \cell{} and LRU neurons. This appendix evaluates on the same benchmark, i.e. the permuted sequential MNIST with and without black pixels, how changing this ratio impacts the performance.

\Cref{fig:seqmnist_hybrid_ratio} shows the mean accuracies obtained for different ratios. While having too many neurons from one type tends to slightly degrade the performance, the intermediate values leads to similar results. This suggests that the performance of hybrid models is not very sensitive to the ratio between the two types of layers, at least for this task.

\section{\texorpdfstring{Making the BMRU parameter $\alpha$ input-dependent}{Making the BMRU parameter ⍺ input-dependent}}
\label[appendix]{app:alpha_bmru}

\begin{figure}[tbp]
    \centering
    \begin{subfigure}[t]{.49\textwidth}
        \centering
        \includegraphics{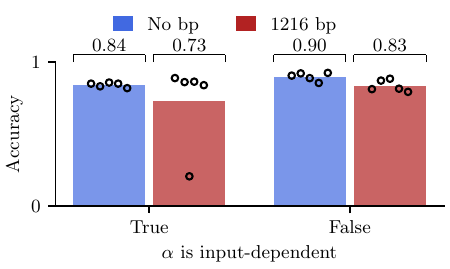}
        \caption{\cell{} alone}
        \label{fig:seqmnist_sbc_alpha}
    \end{subfigure}
    \begin{subfigure}[t]{.49\textwidth}
        \centering
        \includegraphics{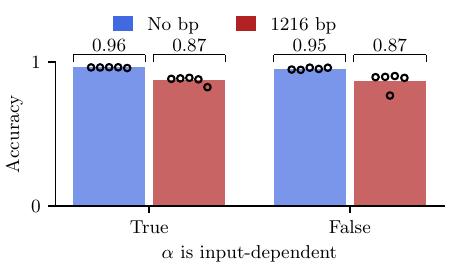}
        \caption{Hybrid model}
        \label{fig:seqmnist_both_alpha}
    \end{subfigure}
    \caption{\textbf{Impact of an input-dependent $\alpha$ on the performance of BMRU.} Accuracies obtained by \cell{} and hybrid models on the permuted sequential MNIST, with or without black pixels added at the end of the sequences, by considering an input-dependent or input-independent $\alpha$ in the equations of BMRU. All models have two recurrent layers with a size of $256$.}
    \label{fig:seqmnist_alpha_bmru}
\end{figure}

The parameter $\alpha$ is introduced in \cref{eq:approx} as the value of the stable state towards which the hysteretic system emulated by BMRU converges. This value is independent on the inputs, which is why $\alpha$ is a learned parameter in BMRU and not a function of the current input. However, from a ML perspective, making it dependent of the current input could improve BMRU's ability to encode information from the specific timestep. This appendix evaluates how adding such dependency can improve or impede performance of BMRU and hybrid models on the permuted sequential MNIST, with and without black pixels. Mathematically, the equations of the modified BMRU are:
\begin{subequations}
    \begin{align*}
        \cand_t  & = W_x x_t + b_x,
        \\
        \beta_t  & = |W_{\beta} x_t + b_{\beta}|,
        \\
        z_t      & = H(|\cand_t| - \beta_t),
        \\
        \alpha_t & = W_{\alpha} x_t + b_{\alpha},
        \\
        h_t      & = z_t \odot S(\cand_t) \odot \alpha_t + (1 - z_t) \odot h_{t-1},
    \end{align*}
\end{subequations}

\Cref{fig:seqmnist_alpha_bmru} shows the mean accuracies obtained for both BMRU and hybrid models and for an input-dependent and independent $\alpha$. Adding such dependency seems to deteriorate BMRU performance when it is used alone, while performance is unaffected for the hybrid model. Therefore, it seems that the quantization imposed by the hysteretic behavior does not impede performance in this task. However, this could not be true in tasks where more information is provided at every timestep, in which case  encoding input-dependent values might be more helpful.

\end{document}